\documentclass[journal,transmag]{IEEEtran}
\usepackage{stfloats}
\usepackage{lineno}
\usepackage{times}
\usepackage{epsfig}
\usepackage{graphicx}
\usepackage{amsmath}
\usepackage{amssymb}
\usepackage{subcaption}
\usepackage{booktabs}
\usepackage{bm}
\usepackage{graphicx}
\usepackage{epstopdf}

\usepackage{hyperref}
\begin{document}

\title{Unstructured Road Vanishing Point Detection Using the Convolutional Neural Network and Heatmap Regression}

\author{\IEEEauthorblockN{Yin-Bo Liu,
Ming Zeng,
Qing-Hao Meng}
\IEEEauthorblockA{Institute of Robotics and Autonomous Systems \\ Tianjin Key Laboratory of Process Measurement and Control \\ School of Electrical and Information Engineering\\ Tianjin University,  Tianjin 300072, China}
\thanks{Corresponding author: Ming Zeng, Qing-Hao Meng (email: zengming@tju.edu.cn, qh\_meng@tju.edu.cn).}}

\IEEEtitleabstractindextext{%
\begin{abstract}
Unstructured road vanishing point (VP) detection is a challenging problem, especially in the field of autonomous driving. In this paper, we proposed a novel solution combining the convolutional neural network (CNN) and heatmap regression to detect unstructured road VP.  The proposed algorithm firstly adopts a lightweight backbone, i.e., depthwise convolution modified HRNet, to extract hierarchical features of the unstructured road image. Then, three advanced strategies, i.e., multi-scale supervised learning, heatmap super-resolution, and coordinate regression techniques are utilized to achieve fast and high-precision unstructured road VP detection. The empirical results on Kong's dataset show that our proposed approach enjoys the highest detection accuracy compared with state-of-the-art methods under various conditions in real-time, achieving the highest speed of 33 fps. 

\end{abstract}

}

\maketitle

\IEEEdisplaynontitleabstractindextext
\IEEEpeerreviewmaketitle

\section{Introduction}
Recently, the research of vanishing point (VP) detection has gradually become one of the most popular topics in the field of computer vision. Vanishing point is defined as the point of intersection of the perspective projections of a set of parallel lines in 3D scene onto the image plane \cite{shi2015fast}. Since the VP of the image contains more valuable cues, it has been widely used in many areas, such as camera calibration \cite{zhang2018orbit},
camera distortion correction \cite{zhu2019distortion}, visual place recognition \cite{pei2019ivpr}, lane departure warning (LDW) \cite{yoo2017robust}, and simultaneous localization and mapping (SLAM) \cite{ji2015rgb}. Specifically, in terms of autonomous driving applications, the detection techniques for structured roads with clear markings have been extensively explored in the literature. However, detecting unstructured roads without clear markings is still a challenging problem that is not perfectly solved \cite{haris2018deep}. The VP provides important clues for unstructured road detection. To be concrete, autonomous vehicles can identify the drivable areas according to the location of the VP, which provides early warning of departure from the lane.

Existing road VP detection methods could be divided into two categories: traditional methods and deep-learning-based methods. Traditional approaches for unstructured road VP detection mainly rely on the texture information extracted from the road images. These texture-based traditional methods usually have two shortcomings:  1) The performance of these methods is easily affected by the factors of uneven illumination and image resolution, and thus the detection results are often unstable, resulting in low VP detection accuracy in many cases; 2) Traditional texture detection and follow-up voting processing are very time-consuming,  which cannot meet the high real-time requirements for autonomous driving. In recent years, deep learning technology has made a series of major breakthroughs in many fields, especially for image recognition. There have been several attempts to utilize deep-learning-based strategies to solve road VP detection problems \cite{lee2017vpgnet,choi2019regression}. However, most of the deep-learning-based methods only focused on structured roads. To the best of our knowledge, there is no solution based on deep learning available in the literature on the subject of unstructured road VP detection.

In this paper, we propose a novel heatmap regression method based on multi-scale supervised learning for unstructured road VP detection. The heatmap regression technique can be used to estimate the locations of pixel-level keypoints in the image and works well in 2D human pose estimation applications \cite{newell2016stacked}. However, the conventional heatmap regression techniques only estimate the keypoints on the 1/4 or 1/2 single, coarse-scale of the input image. Therefore, it cannot meet the requirement of high-precision road VP detection in the application of autonomous driving. The proposed method adopts three effective tricks such as multi-scale supervised learning, heatmap super-resolution and coordinate regression to achieve fast and accurate unstructured road VP detection. The experimental results on the public Kong's dataset verify the effectiveness of the proposed algorithm.  

The main contributions are as follows:
\begin{itemize}
\item To the best of our knowledge, the proposed approach is the first attempt to solve the problem of unstructured road VP detection with the deep CNN architecture. Specifically, it integrates three advanced strategies including improved lightweight backbone design, multi-scale supervised learning, and heatmap super-resolution, which make the proposed algorithm have advantages of high accuracy and rapidity.
\item Our approach can run at 33 fps on an RTX 2080 Ti GPU, which is several times faster than state-of-the-art methods. 
\item In order to evaluate the performance of different algorithms more accurately, we have constructed a manually labeled training dataset including 5,355 images of unstructured roads.
\end{itemize}

The remainder of this paper is organized as follows. We first review some relevant works in Section \ref{sec2}. In Section \ref{sec3}, the proposed algorithm is introduced in detail. Then, we compare the performance of different algorithms in Section \ref{sec4}, and followed by the conclusions in Section \ref{sec5}

\section{Related work}\label{sec2}

We firstly introduce some traditional algorithms about road VP detection. Secondly, a brief review of the state-of-the-art algorithms in the field of heatmap regression is given.

\subsection{Road VP detection}
The traditional methods found in the literature could be divided into three categories, i.e., edge-based detection methods, region-based detection methods, and texture-based detection methods. Among them, edge-based and region-based detection approaches are commonly used for detecting the VPs in the structured roads. The edge-based detection algorithms estimate the VPs based on the information of road edges and contours, e.g., the spline model proposed by Wang et al. \cite{wang2000lane}, the cascade Hough transform model presented by Tuytelaars et al. \cite{tuytelaars1998cascaded}. The B-snake based lane model proposed by Wang et al. \cite{wang2004lane} can describe a wider range of road structures due to the powerful ability of B-Spline, which can form any arbitrary shape. Instead of using hundreds of pixels, Ebrahimpour et al. \cite{ebrahimpour2012vanishing} import the minimum information (only two pixels) to the Hough space to form a line, which greatly reduces the complexity of the algorithm. The region-based methods locate the VPs by analyzing the similar structures and repeating patterns in the road images. Specifically, Alon et al. \cite{alon2006off} utilize the technique of region classification and geometric projection constraints to detect the road VP. The region features of the self-similarity\cite{kogan2009vanishing} and  road boundaries \cite{wang2017fast} are useful features for road VP detection. In addition, Alvarez et al. \cite{alvarez20103d} utilize 3D scene cues to predict the VP. Although the above-mentioned edge-based and region-based approaches work well on the simple structured road scenes, for the complicated scenes of unstructured roads, the detection performance is usually poor or even completely invalid \cite{kong2012generalizing}. The main reasons for unsatisfying results are that the unstructured roads often do not have clear lane and road boundaries, and there are many disturbances such as tire or snow tracks.

In traditional texture-based methods, a large number of textures are firstly extracted from the road image, and then the strategy of accumulating pixel voting is adopted to detect the road VP. For example, Rasmussen et al. \cite{rasmussen2008roadcompass} use Gabor wavelet filters for texture analysis and the Hough-style voting strategy for VP detection. Kong et al. \cite{kong2010general} employe 5-scale, 36-orientation Gabor filters to extract the textures, and then utilize the adaptive soft voting scheme for VP detection. To accelerate the speed of VP detection, Moghadam et al. \cite{moghadam2011fast} only use four Gabor filters in the process of feature extraction. In order to obtain better textures, Kong et al. \cite{kong2012generalizing} replace conventional Gabor filters with generalized Laplacian of Gaussian (gLoG) filters. Yang et al. \cite{yang2016fast} adopt a Weber local descriptor to obtain salient representative texture and orientation information of the road. Shi et al. \cite{shi2015fast} use a particle filter for reducing the misidentification chances and computational complexity, and a soft voting scheme for VP detection. Generally speaking, the texture-based methods can detect the VP for both the structured and unstructured roads. However, these methods also have common shortcomings: 1) Texture detection performance strongly depends on image quality. For the image with uneven illumination or poor definition, the extracted textures are usually not good, which has a great negative impact on the accuracy of VP detection; 2) The process of cumulative pixel voting to predict the VP is time-consuming, which cannot be directly used for the scenarios where real-time requirements are high, e.g., the scenario of automatic driving.

\subsection{Heatmap regression for keypoint detection }

In recent years, the research of deep-learning-based heatmap regression has attracted considerable attention in the field of image keypoint detection. The CNN based heatmap regression scheme is firstly applied in human pose estimation \cite{wei2016convolutional}. The empirical results show that it can accurately predict the probability of human joints with pixel resolution, which outperforms most of traditional keypoint detection methods. Subsequently, the CNN based heatmap regression methods are successfully introduced to other application areas, such as general target recognition \cite{law2018cornernet}, layout detection \cite{lee2017roomnet} and target tracking \cite{ning2017spatially}. In the early heatmap regression networks, the conventional Resnet module \cite{resnet} is widely used as the backbone, but the detection accuracy is not satisfactory. Although recently modified versions, such as stacked hourglass network (Hourglass) \cite{newell2016stacked} and high-resolution representation network (HRNet) \cite{sun2019deep} greatly improve the accuracy of keypoint detection, they still have a common shortcoming of long prediction time, which is not suitable for high-speed scenarios.

\section{Methodology}\label{sec3}

\begin{figure*}[t]
\centering
\includegraphics[scale=0.65]{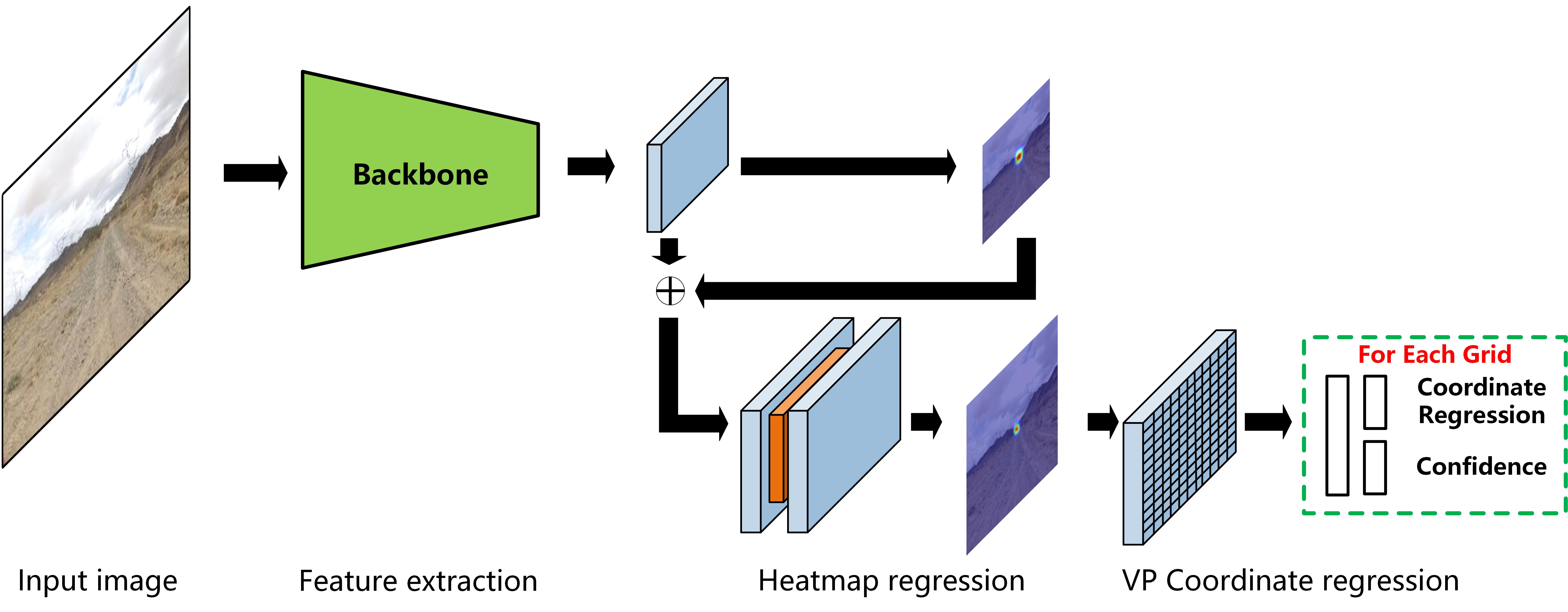}
\caption{Illustration of our proposed network architecture. Firstly, we adopt the depthwise convolution \cite{mobilenetv2} modified HRNet as the backbone to extract hierarchical features of the input image.  Then, we combine 1/4 and 1/2 scale heatmaps for multi-scale supervised learning. Finally, the coordinate regression is employed and then directly output high-precision VP coordinates. }%
\label{fig:label1}
\end{figure*}

\begin{figure*}[htbp]
\centering
\includegraphics[scale=0.8]{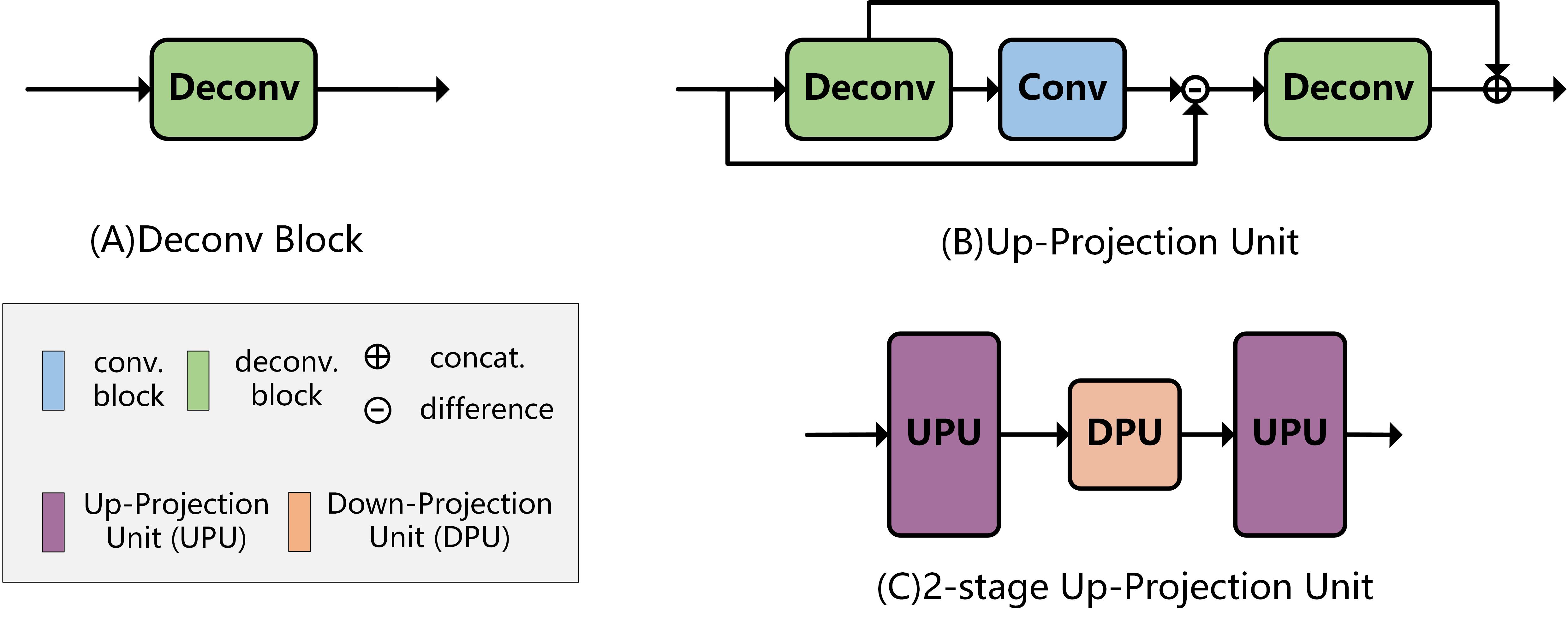}
\caption{Illustration of 3 types from 1/4 scale upsamples to 1/2 scale heatmap. (A) represents a layer of deconvolution + BatchNorm + ReLU. (B) is the up-projection unit used in this paper. (C) is a 2-level up-projection model.}%
\label{fig:label6}
\end{figure*}

Previous empirical results show that the CNN based heatmap expression is an advanced technology for image keypoint detection and it can locate the keypoint with pixel-level resolution. At present, this methodology has achieved good results in the application of 2D human pose estimation. The road VP can be regarded as a special kind of keypoints. Therefore, we can also use this advanced keypoint detection technology to deal with the challenging problem of unstructured road VP detection. Concretely, the modified HRNet is firstly utilized as the backbone to extract the image features. Secondly, multi-scale heatmap supervised learning is employed to obtain more accurate keypoint (Vanishing Point) estimation. Finally,  high-precision VP coordinates are obtained using the strategy of coordinate regression. The specific network architecture is shown in Fig. \ref{fig:label1}.
\subsection{Multi-Scale Supervision}
The latest research suggests that multi-scale supervised learning is an effective way to obtain accurate heatmaps \cite{sun2019deep}. Multi-scale heatmap supervised learning refers to fusing two or more scale heatmaps for keypoint detection. Generally, in traditional heatmap regression methods, the analyzing scale of the heatmap is 1/4 of the input image and the corresponding resolution can meet the requirements of most ordinary keypoint detection tasks. However, for the task of road VP detection, 4 times error amplification is hard to accept. To this end, we obtain a more accurate and higher-resolution 1/2 scale heatmap from the coarse-grained 1/4 scale heatmap using the back-projection-based super-resolution technique \cite{haris2018deep}.

The super-resolution module used in our proposed algorithm is an up-projection unit (UPU), as shown in Fig. \ref{fig:label6}. Each sub-module of Deconv consists of a $5 \times 5$ deconvolution operation followed by BatchNorm and ReLU operations, and each sub-module of Conv is composed by a $3 \times 3$ convolution operation followed by BatchNorm and ReLU operations. As suggested in higherHRNet \cite{cheng2019bottom}, the input of our upsample module is the concatenation of the feature maps and the predicted  1/4 scale heatmap. In the subsequent ablation study, we will systematically compare the performance difference using the super-resolution module with other upsampling modules.
\subsection{Coordinate Regression}
In order to obtain the coordinates of a keypoint from the heatmap, there are two widely used traditional methods, i.e., extracting the coordinates of the maximum point in the heatmap or estimating the keypoint position through a Gaussian distribution. The traditional methods have two limitations: 1) Since the resolution of the heatmap is generally 1/2 or 1/4 scale of the input image, the estimated error is amplified 2 times or 4 times accordingly; 2) Estimating the coordinates based on Gaussian distribution requires additional calculations which affect the real-time performance of the algorithm. To overcome the shortcomings of traditional methods, the proposed algorithm introduces a coordinate regression module to directly output accurate VP coordinates.

\begin{figure*}[!h]
\centering
\includegraphics[scale=0.8]{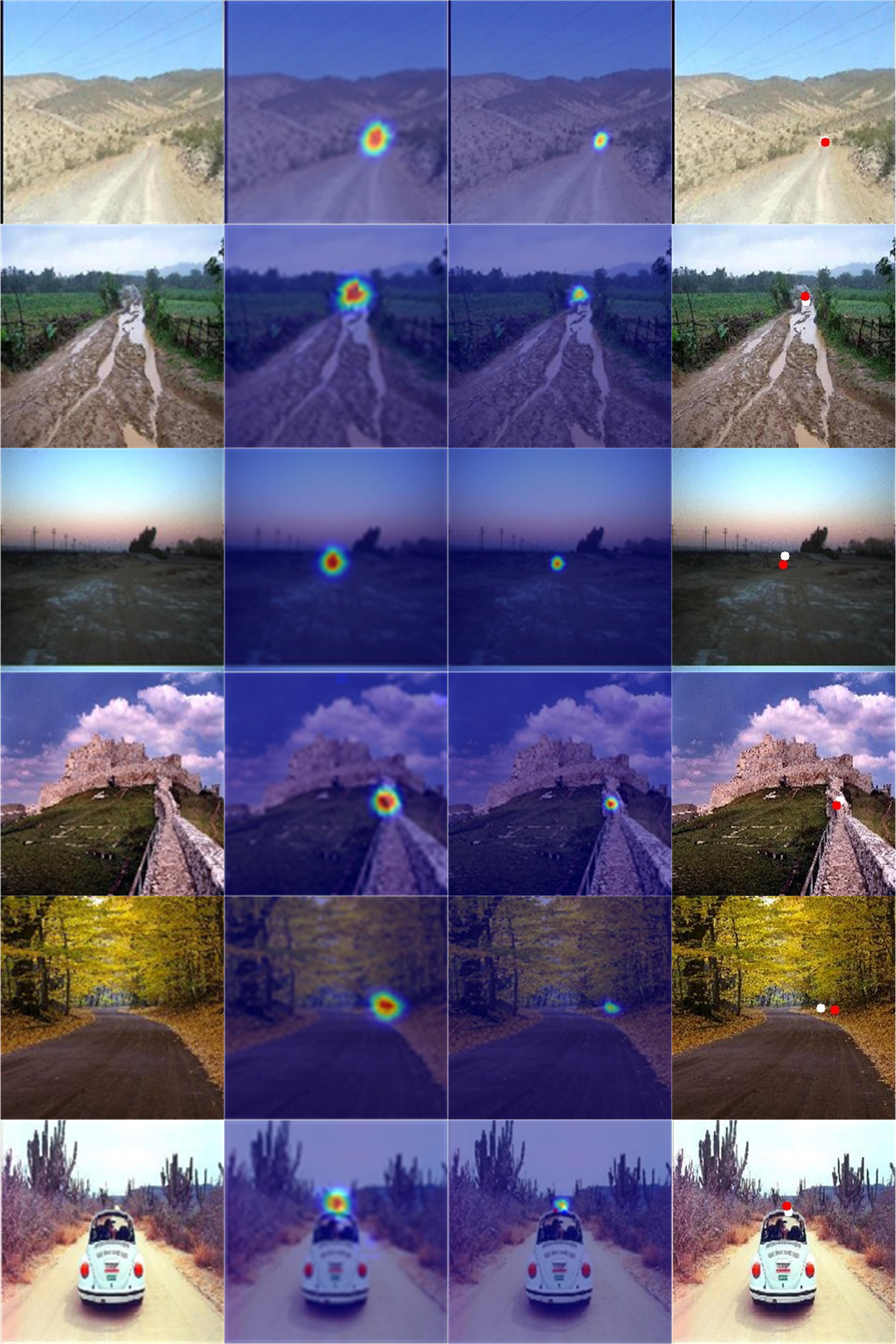}
\caption{The above figure shows six of the detection results of our method. From left to right: input image, 1/4 scale VP estimation heatmap, 1/2 scale VP estimation heatmap, VP coordinates determined by coordinate regression (the red dot represents the VP position detected by the proposed algorithm, and the white dot is the ground truth).}%
\label{fig:label2}
\end{figure*}

The coordinate regression module used in our approach is inherited from YOLO. As proven in YOLO v2 \cite{redmon2017yolo9000}, the prediction of the offset to the cell is a stabilized and accurate way in 2D coordinate regression. Thus, the predicted point $(V_{x},V_{y})$ is defined as 
\begin{gather}
V_{x}=f(x)+c_{x}, \\
V_{y}=f(y)+c_{y},
\end{gather}

\noindent where $f(\cdot)$ is a sigmoid function of the road VP. $(c_{x},c_{y})$ is the coordinate of the top-left corner of the associated grid cell.
\subsection{Loss Function}
To train our complete network, we minimize the following loss function.
\begin{equation}
L_{vp}=\lambda_{coord}l_{coord}+\lambda_{conf}l_{conf}+\lambda_{h}(l_{h1}+l_{h2}),
\end{equation}

\noindent where $l_{coord}, l_{h1}, l_{h2}$ and $l_{conf}$  denote the coordinate losses of the VP, the low resolution heatmap loss, the high resolution heatmap loss and the confidence loss, respectively. We use the mean-squared error for the VP heatmap loss, while as suggested in YOLO, the confidence and VP coordinates are predicted through logistic regression. For the cells that do not contain the VP, we set $ \lambda_{conf} $ to 0.5, and for the cell that contains VP, we set $ \lambda_{conf} $ to 1. $ \lambda_{coord} $ and $ \lambda_{h} $ are used for balancing the two training factors, i.e., the heatmap accuracy and coordinate accuracy. Here, we set $ \lambda_{h} $ to 1 and $ \lambda_{coord} $ to 2.

\subsection{Implementation Details}
We implemented our method in Python using Pytorch 1.3 and CUDA 10 and ran it on an i7-8700K@3.7GHz with dual NVIDIA RTX 2080 Ti. We used our unstructured road vanishing point (URVP) dataset as the training dataset, which contains 5,355 images, and Kong's public dataset as the test dataset. All input images were reshaped to 320 $ \times $ 320 for training.  We applied a Gaussian kernel with the same standard deviation (std = 3 by default) to all these ground truth heatmaps. We used stochastic gradient descent (SGD) for optimization and started with a learning rate of 0.001 for the backbone and 0.01 for the rest of the network. We divided the learning rate by 10 every 20 epochs, with a momentum of 0.9. We also adopted data augmentation with random flip and image rotations.

\section{Experiments}\label{sec4}

In this section, we first briefly introduce the construction of the training dateset, a.k.a. URVP dataset, and then illustrate a comprehensive performance study of our proposed method and other state-of-the-art algorithms using Kong's public dataset. Subsequently, we quantitatively analyze the influence of each part of the model on performance through an ablation study.

\begin{figure*}[htbp]
\centering
\includegraphics[scale=0.50]{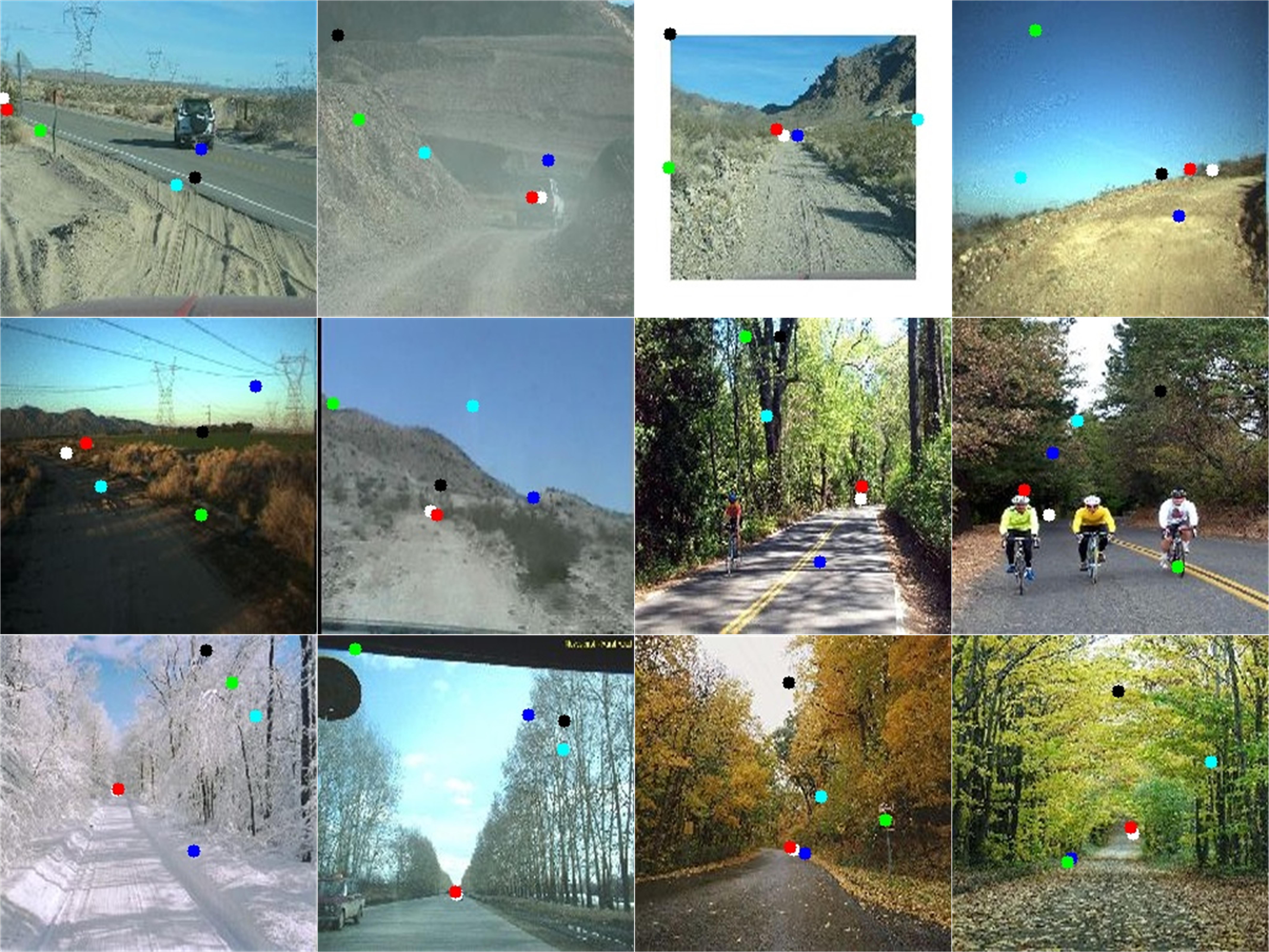}
\caption{Experimental results on some test images from Kong's dataset. Red dot denotes the results of our algorithm, black ones Kong (Gabor), light blue ones Moghadam, green ones Yang, blue ones Kong (gLoG), and white ones the ground truth.}%
\label{fig:label3}
\end{figure*}

\subsection{Dataset construction}
Currently, there are only two public databases available to evaluate the performance of different algorithms of unstructured road VP detection, i.e., Kong's dataset (containing 1003 images) and Moghadam's dataset (500 images). In view of the fact that the total number of images is very small, these two public datasets are mainly used for algorithm testing. In other words, there are not enough labeled images for deep network training. To this end, we utilized the tools of Flickr and Google Image to build a new training dataset, namely URVP. Specifically, we first collected more than 10,000 unstructured road images using related keywords such as vanishing point, road, and unstructured road. Subsequently, after data cleaning and manual labeling, 5,355 labeled images were finally obtained as the training dataset.

\subsection{Metrics}
We adopt the normalized Euclidean distance suggested in \cite{moghadam2011fast} to measure the estimation error between the detected VP and the ground truth manually determined through the perspective of human perception. The normalized Euclidean distance is defined as:
\begin{equation}
NormDist=\frac{\lVert P_{g}-P_{v} \rVert}{Diag(I)},
\end{equation}
where $P_{g}$ and $P_{v}$ denote the ground truth of the VP and the estimated VP, respectively. $Diag(I)$ is the length of the diagonal of the input image. The closer the $NormDist$ is to 0, the closer the estimated VP is to the ground truth. The $NormDist$ greater than 0.1 is set to 0.1, which is considered to be a failure of the corresponding method.

\subsection{Comparisons}

\begin{figure}[htbp]
\centering
\includegraphics[scale=0.6]{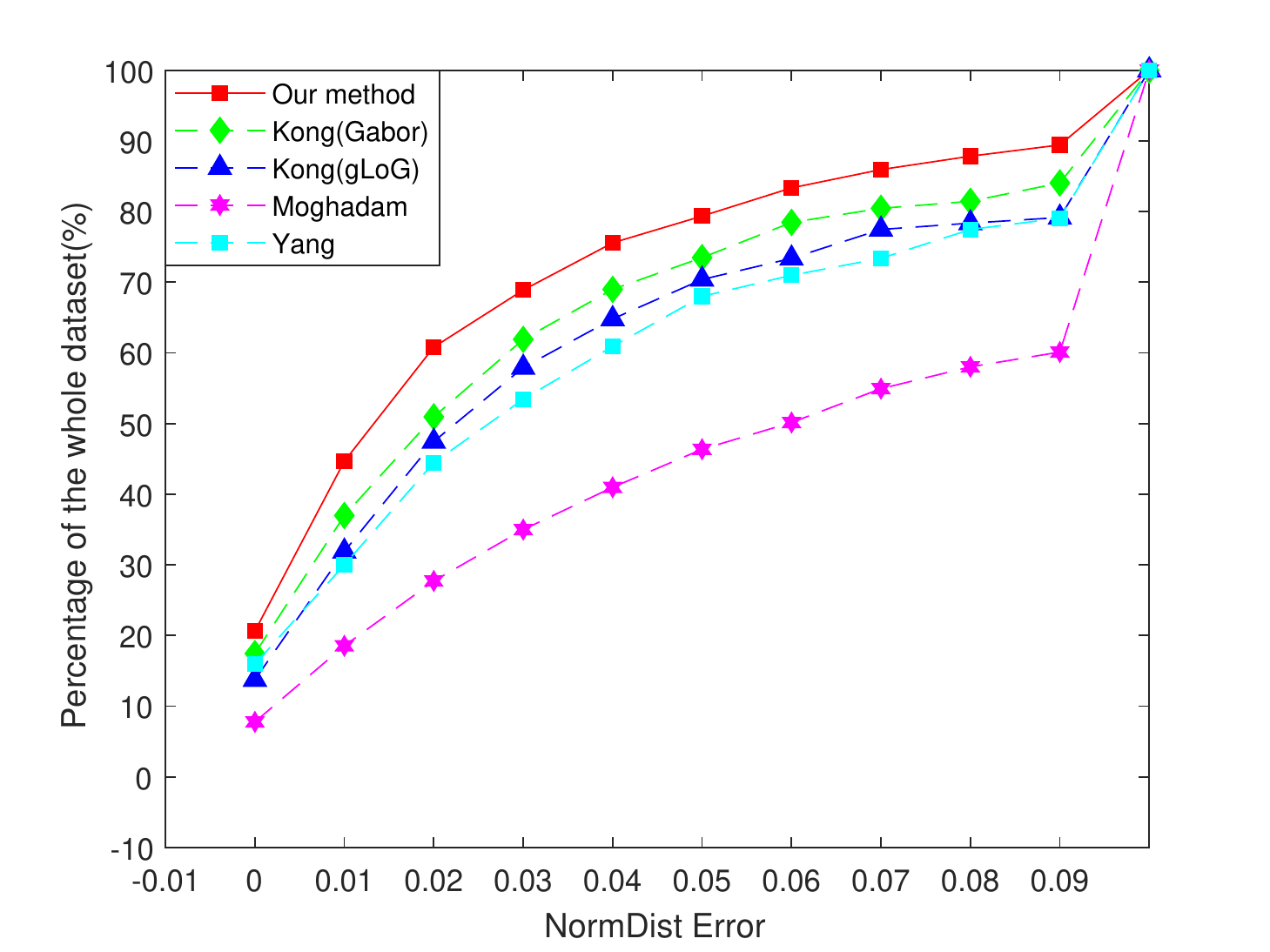}
\caption{Comparison results of accumulated error distribution of different VP detection algorithms in Kong's dataset. On the x-axis, 0 stands for $NormDist$ in [0,0.01), 0.01  stands for $NormDist$ in [0.01, 0.02)..., and 0.1 represents $NormDist$ in [0.1,1].}%
\label{fig:label4}
\end{figure}
Fig. \ref{fig:label2} shows the test results of the proposed algorithm on unstructured road images. From left to right are input image, 1/4 scale output heatmap, 1/2 scale output heatmap and VP coordinates detected by coordinate regression. The white dot represents the ground truth of VP, and the red dot stands for the predicted VP coordinate. It is obvious that from left to right, the possible range of the detected VP is gradually reduced and the proposed method achieves good results on the unstructured roads.

In view of the fact that the Moghadam's dataset has small number of testing images (only 500 images) and 248 images are the same as those of the Kong's dataset, we only compared the proposed algorithm with four state-of-the-art methods, i.e, Kong (Gabor) \cite{kong2010general}, Kong (gLoG) \cite{kong2012generalizing}, Moghadam \cite{moghadam2011fast}, Yang \cite{yang2016fast} on the Kong's dataset (1003 pictures). Fig. \ref{fig:label3} shows some VP detection examples, in which white dots denote the ground truth results, red ones are results of our method, black ones are results of Kong (Gabor), light blue ones are Moghadam's results, green ones are Yang's results, blue ones are results of Kong (gLoG). Obviously, our proposed algorithm is more robust and accurate than the existing state-of-the-art methods.

Furthermore, we quantitatively evaluate the performance of our method with normalized Euclidean distance. The test results show that the proposed detection algorithm outperforms other existing road VP detection algorithms. More specifically, the proposed method has the highest percentage of $NormDist$ error in [0, 0.1) compared with the four existing VP detection algorithms, as shown in Fig. \ref{fig:label4}.  Our method can detect 207 images with $NormDist$ error less than 0.01, while the number of images detected by Kong (Gabor), Yang, Kong (gLoG) and Moghadam is 175, 160, 138 and 78, respectively. For cases with large detection errors ($NormDist \geq 0.1$), the proposed algorithm only contains 103 images, while the comparison methods are Kong (Gabor) (160), Kong (gLoG) (209), Yang (210) and Moghadam (400). In addition, Table. \ref{tab1} shows the statistical results of the mean runtime of CPU for different methods. It can be seen that our method is much faster than the counterparts of Kong (Gabor), Kong (gLoG), Yang and Moghadam's algorithms.

\begin{table}[h]
\centering
\caption{Comparison results of mean error and mean running time for different approaches. Here, we only compared the CPU running times since the Kong (Gabor), Kong (gLoG), Yang and Moghadam's methods are  non-CNN methods which can not run on the GPU.}
\scalebox{1.2}{
\begin{tabular}{ccc}
\hline
Methods     & Mean error & CPU Running Time (s) \\ \hline
Kong (Gabor) & 0.040639   & 20.1021             \\
Kong (gLoG)  & 0.051556   & 21.213              \\
Moghadam    & 0.063407   & 0.2423              \\
Yang        & 0.045931   & 0.752               \\
Proposed    & \textbf{0.034875}   & \textbf{0.2024}              \\ \hline
\end{tabular}}
\label{tab1}
\end{table}

\subsection{Ablation Study}
The multi-scale supervision, upsampling module, and coordinate regression are three important branches in our model. Therefore, we will quantitatively analyze the impact of these branches on the performance of the detection network.
\subsubsection{Backbone}

We systematically tested the effects of different backbones, i.e., Hourglass ($Stack = 4$) (Hg4) and HRNet ($W = 48$) (HRNet-48), on accuracy and detection speed of the model. Table \ref{tab2} shows the accuracy, GPU and CPU runtime with different backbones. To be concrete, using the depthwise convolution modified HRNet-48 (HRNet-48-M), the number of images with a detection error $NormDist \textless 0.01$ is 207, which is higher than the models using Hg4 (192) or HRNet-48 (205). For cases with large detection errors ($NormDist$ is equal to or greater than 0.1),  unsatisfying results for the model using HRNet-48-M is 103, which is smaller than those of models using HRNet-48 (115) and Hg4 (113). Moreover, the GPU estimated speed of the model using HRNet-48-M is 33 fps, which is faster than those of models using Hg4 (23 fps) and HRNet-48 (29 fps). Obviously, the model using HRNet-48-M has two advantages, i.e., high speed and high accuracy over the models with Hg4 and HRNet48.

\subsubsection{Multi-scale Supervision}

\begin{figure}[htbp]
\centering
\includegraphics[scale=0.6]{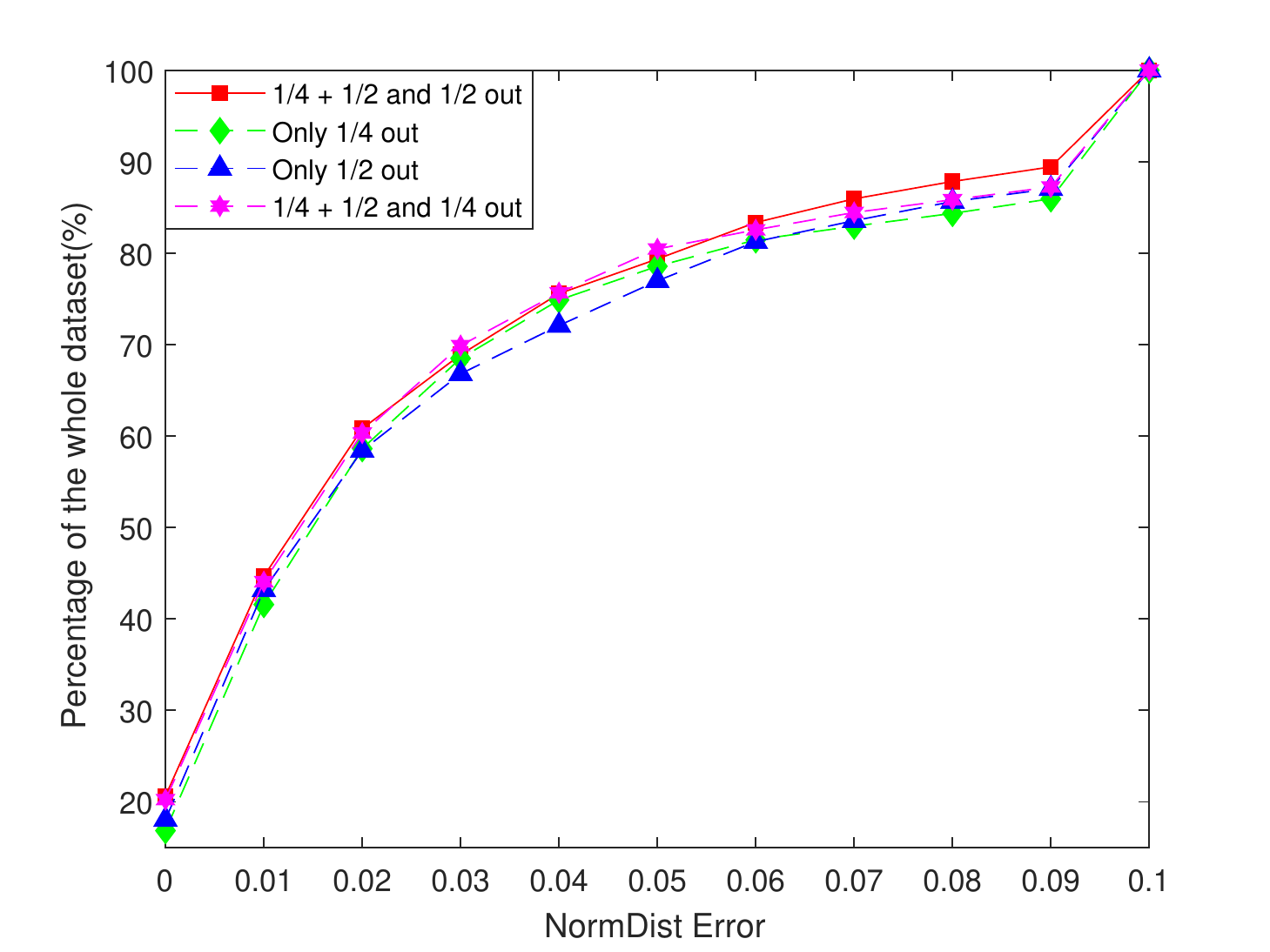}
\caption{Ablation study of multi-scale supervision. Illustration of the output of a single 1/4 scale and 1/2 scale supervised learning coordinate regression, 1/2 and 1/4 scale fusion learning and 1/4 scale coordinate regression, 1/2 and 1/4 scale fusion learning and 1/2 scale coordinate regression.}%
\label{fig:label5}
\end{figure}

The influence of multi-scale supervised learning on detection performance is shown in Fig. \ref{fig:label5}. When only the 1/2 scale supervised learning model is adopted, the number of images with detection error $NormDist$ less than 0.01 is 181. When only 1/4 scale supervised learning model is applied, the number of images satisfying the accuracy is 169. When 1/2 and 1/4 scales are fused to implement multi-scale supervised learning and 1/4 scale of the input image is used to make coordinate regression, the number of images satisfying $NormDist \textless 0.01$ is 203. When we adopt multi-scale supervised learning and 1/2 scale of the input image for coordinate regression, we can detect 207 images with $NormDist \textless 0.01$. Obviously, the last one is the best choice.

\subsubsection{Upsampling Module}
We select three different upsampling modules (Fig. \ref{fig:label6}) and measure the impact of different module selection on the performance of VP detection. The results are shown in Table \ref{tab3}. When the deconvolution module (A) is adopted, the mean error of $NormDist$ is 0.035541. If we select the  2-stage Up-Projection Unit (C), the mean error of $NormDist$ is 0.034954. And when using the Up-Projection Unit (B), it can achieve the smallest mean error (0.034875). Therefore, the VP detection network selects the Up-Projection Unit for the upsampling operation.

\begin{table*}[t]
\centering
\caption{Ablation study of different backbones. GPU-speed and CPU-speed represent the number of frames that different backbones run on the GPU and CPU, respectively.}%
\scalebox{1.2}{
\begin{tabular}{ccccc}
\hline
Backbones          & \begin{tabular}[c]{@{}c@{}}Number of images with\\ NormDist error $\textless 0.01 $ \end{tabular} & \begin{tabular}[c]{@{}c@{}}Number of images with\\ NormDist error $\geq 0.1$ \end{tabular} & GPU-speed & CPU-speed \\ \hline
Hg4 & 192                  & 113                &23.04 fps    &2.02 fps         \\
HRNet-48              & 205                  & 115                &29.15 fps    &2.90 fps         \\
HRNet-48-M       & \textbf{207}                  & \textbf{106}                &\textbf{33.05 fps}    &\textbf{4.94 fps}         \\ \hline
\end{tabular}}
\label{tab2}
\end{table*}

\begin{table}[]
\centering
\caption{Ablation study of different upsampling modules.}
\begin{tabular}{c|c|c|c|c}
Upsampling & \begin{tabular}[c]{@{}c@{}}w/ Deconv\\ block\end{tabular} & \begin{tabular}[c]{@{}c@{}}w/ UPU \end{tabular} & \begin{tabular}[c]{@{}c@{}}w/ 2-stage \\ UPU\end{tabular} & mean error \\ \hline
DB                                                                &\checkmark     &                &                       &0.035541            \\
UPU               &               &\checkmark      &                       &\textbf{0.034875}            \\
2-stage UPU     &               &                &\checkmark             &0.034954
\end{tabular}
\label{tab3}
\end{table}

\subsubsection{Coordinate Regression}
\begin{table}[]
\centering
\caption{Ablation study of different regression components such as single heatmap regression, multi-scale heatmap regression, and coordinate regression.}%
\begin{tabular}{@{}c|c|c|c|c@{}}
Type & \begin{tabular}[c]{@{}c@{}}w/ heatmap\\ regression\end{tabular} & \begin{tabular}[c]{@{}c@{}}w/ multi-scale\\ regression\end{tabular} & \begin{tabular}[c]{@{}c@{}}w/ coordinate\\ regression\end{tabular} & mean error \\ \midrule
a    &\checkmark                 &                           &               &0.035416            \\
b    &\checkmark                 &\checkmark                 &               &0.035152            \\
c    &\checkmark                 &\checkmark                 &\checkmark     &\textbf{0.034875}
\end{tabular}
\label{tab4}
\end{table}
Finally, we tested the influence of different coordinate regression selections on detection performance. The results are shown in Table \ref{tab4}.  When we only use heatmap regression, the mean error of $NormDist$ is 0.035416. When using a combination of heatmap + multi-scale regression, the mean error of $NormDist$ is 0.035152. When using the strategy of heatmap + multi-scale regression + coordinate regression, the mean error of $NormDist$ is 0.034875. Therefore, we select the last strategy to estimate the coordinates of the road VP.

\section{Conclusion}\label{sec5}

Quickly and accurately detecting the vanishing point (VP) in the unstructured road image is significantly important for autonomous driving. In this paper, a CNN based heatmap regression solution for detecting the unstructured road VP is proposed. The modified lightweight backbone, i.e., depthwise convolution modified HRNet, improves the detection speed to 33 fps on an RTX 2080 Ti GPU, which is several times faster than that of the state-of-the-art algorithms. In addition, three useful tricks such as multi-scale supervised learning, heatmap super-resolution, and coordinate regression are utilized, which make the proposed approach achieve the highest detection accuracy compared with the recent existing methods. In the future, we plan to utilize the proposed VP detection technique to carry out the research of multi-task learning for accurate lane detection. Our code and constructed URVP dataset will be made publicly available for the sake of reproducibility

\section*{Acknowledgment}
This work is supported by the National Natural Science Foundation of China (No. 61573253), and National Key R\&D Program of China under Grant No. 2017YFC0306200.
\bibliography{ms}
\end{document}